# An Effective Two-Phase Genetic Algorithm for Solving the Resource Constrained Project Scheduling Problem (RCPSP)

D. Sun and S. Zhou

## Abstract

This note presents a simple and effective variation of genetic algorithm (**GA**) for solving **RCPSP**, denoted as 2-Phase Genetic Algorithm (**2PGA**). The **2PGA** implements **GA** parent selection in two phases: Phase-1 includes the best current solutions in the parent pool, and Phase-2 excludes the best current solutions from the parent pool. The **2PGA** carries out the **GA** evolution by alternating the two phases iteratively. In exploring a solution space, the Phase-1 emphasizes intensification in current neighborhood, while the Phase-2 emphasizes diversification to escape local traps. The **2PGA** was tested on the standard benchmark problems in **PSPLIB**, the results have shown that the algorithm is effective and has improved some of the best heuristic solutions.

Key words: resource-constrained project scheduling problem, RCPSP, genetic algorithm, selection pressure, metaheuristics, optimization.

## 1 Introduction

The Resource-Constrained Project Scheduling Problem (**RCPSP**) is a well-known classic optimization problem. Its purpose is to schedule a set of project activities, subject to precedence and renewable resource capacities constraints, with the goal of minimizing the overall project total completion time (make-span)

The **RCPSP** is known to be **NP-hard**, meaning that for large instances, finding the optimal solution in a reasonable amount of time is infeasible. Therefore, **Metaheuristic Algorithms,** such as genetic algorithm (**GA**), tabu search, simulated annealing, scatter search, etc., are commonly used to solve the problem [4, 7, 8, 12, 23]. Among the **Metaheuristic Algorithms**, **GA** is the mostly studied approach.

The **GA** is a population-based algorithm. It is based on the **survival of the fittest** principle, and imitates the biologic evolution through 1) parent **selections**, 2) **crossover** to produce the next generation, and 3) **mutations**. The **GA** select good solutions from a large diversified solution population it generated, and it has been shown to be effective in solving a variety of discrete optimization problems including **RCPSP**.

A variety of **GA**-based approaches for solving **RCPSP** have been proposed, such as presented in [2, 6, 7, 11, 16, 18, 22, 25], and many others. A hybrid approach is also a very active area, e.g., **GA + local-search** or **GA + exact-methods**. The goal is to leverage the global exploration of **GA** with the local exploitation of other methods, to build more powerful and robust algorithms [7, 11]. In this note, we focus on a basic but less studied area: the parent **selection** for next generation in **GA**.

A simple variation of genetic algorithm is presented, called **2PGA,** for solving **RCPSP**. The **2PGA** both reinforces and relaxes the **survival of the fittest** principle. The parent **selection** is implemented in two phases: Phase-1includes the best current solutions into the parent pool, and Phase-2 excludes the best current solutions from the parent pool. The 2 phases alternate through the iteration of **2PGA**. By alternating the inclusion and exclusion of the current best solution, the **2PGA** balances the intensification and diversification in searching the solution space. The testing results on **PSPLIB** [17] have shown that this approach is effective**.**

## 2 Resource Constrained Project Scheduling Problem

**Activities**

A project needs to complete a set A = {1*, ..., N*} activities, each activity is represented as an integer. Dummy activities 0 and $N$+1 are added to represent the project's start and end, respectively. There are precedential relations amount the activities, each activity can only start after all its predecessors are finished. The immediate predecessors of activity $j$ are denoted by set $P_j$. The dummy activity 0 is the predecessors all activities and dummy activity *N+1* has all activities as its predecessors.

**Resources**

Activities need resources to process. There are $K$ renewable resource types, resources are denoted as set R = {$R_1$,..., $R_K$}. Resource $R_k$ has a limited capacity of $RC_k$. The resource requirement of activity $j$ on $R_k$ *is* denoted as $r_{j,k}$. The uninterrupted processing time of activity $j$ is denoted as $t_j$. Dummy activities *0* an *N+1* don't have resource requirement. The start time of activity j on resource k is denoted as $s_j$, and the finish time $f_j = s_j + t_j$. The processing time of dummy activities 0 and N+1 *are* $t_0 = t_{N+1} = 0$.

**RCPSP**

The goal for solving RCPSP is to minimize the time to finish all activities, which is known as ***make-span***, subject to the activity precedence and resource capacity constraints. A project schedule is a set start/finishing times $s_j / f_j$, *j=0, ..., N+1*. A **RCPSP** can be formulated as follows.

$$\begin{aligned}
&\min\ (f_{N+1}) \\
&\text{s.t.} \\
&\textstyle\sum_{j\in A_k} r_{j,k} < RC_k,\ \forall\ k, R_k \in R \\
&s_j\ \geq f_{i\in P_j},\ \forall\ j\in A
\end{aligned}$$

where $A_k$ is the set of activities that are processed on resource k at the same time; $RC_K$ is the capacity of resource $R_k$; $P_j$ is the set of immediate predecessors of activity j, $R$ is the set of all resources, $A$ *is* the set of all activities.

## 3 Two Phase Genetic Algorithm (2PGA)

**Schedule Representation**

An *activity-list* approach [12] is used in **2PGA** to represent an individual schedule. An *activity-list* is a permutation of all activities in a precedence feasible order and each activity is represented as an integer *j* ∈ {*0, ..., N+1*}. *activity-lists* are encoded schedules and correspond to chromosomes in genetic operations, which are the basic units to perform crossovers, mutations, and to form populations.

**Serial Schedule Generation Scheme (SSGS)**

An *activity list* (AL) is an encoding of a schedule. An AL is converted to a project schedule using SSGS [12] that is a decoding process. Assume an AL is implemented as an array denoted as A, SSGS iterates from $A[0]$ to $A[N+1]$ sequentially, note that the order of activities in AL satisfies precedence constraints. At step $i$ of SSGS, assume $A[i]$ contains activity $j$, $j$ needs to wait for two conditions to start: 1) the finish of all its predecessors $(P_j)$; and 2) the available capacities can satisfy its resource requirements. By iterating through an AL, SSGS assigns start/finish times to all activities in the AL, $A[N+1] = \text{activity N} + 1$, is a dummy activity, its start/finish time $s_{N+1} = f_{N+1}$ is the project's ***make-span***.

**Crossover**

The solution space of **2PGA** contains all precedence feasible activity permutations represented by ALs (*activity lists*). For generating new permutations, a ***two-point crossover*** operation is used in **2PGA** to mix two existing ALs (parents) to produce two new ALs (children). Denote the two parents ALs as F and M, and the two children ALs as D an S, the two-point crossover operation to produce D is as follows:

1. Generate 2 random numbers $r_1$ and $r_2$, $0 < r_1 \leq r_2 \leq N$
2. Set $D[i] = M[i], 0 < i \leq r_1$
3. Set $D[i] = F[j], r_1 < i \leq r_2, 0 < j \leq N, j\ is\ lowest\ index$,
   $F[j]$ not in D
4. Set $D[i] = M[j], r_2 < i \leq N,\ r_1 < j \leq N, j\ is\ lowest\ index$,
   $M[j]$ not in D

By switching M and F, another child S can be produced in the same manner.

It has been proved that the crossover operation keeps the *children ALs* precedence feasible [12].

**Mutation**

A mutation operation changes some activities' order in *an AL.* If a mutation in activity permutation violates the precedence feasibility, a repair procedure is then carried out for correction.

Mutations in an *activity list* can be in several forms: 1) move one activity to a different position; 2) exchange two activities' positions; 3) move several activities simultaneously to different positions; and 4) a combination of the above.

The repair procedure works by iterate through the activities that violate the precedence, and move each of them backward to a feasible position.

**Initial Population Generation**

The activities and their precedence relations correspond a directed acyclic graph (DAG), with nodes as activities and directed edges as their precedence relations. Generating a precedence feasible activity list corresponds to a *topological sort* on the DAG. The result of the *topological sort* establishes a partial order for activities. The precedence feasible *activity list* is generated using the following ***topological sort***.

The **eligible activities** are those that either don't have predecessors or their predecessors are already in AL, we keep a set **E** of **eligible activities** in generating a precedence feasible AL through the following steps:

0. Initialize AL size = 0, append the dummy activity 0 to AL,
   Initialize E with eligible activities,
1. Randomly pick an activity in E and append to AL,
2. If some ineligible activities become eligible, add them to set E,
3. Repeat step 1 and 2, until E becomes empty, then append the dummy activity *N+1* to AL.

The generated ALs will be called ***individuals*** in the following description, where each ***individual*** is a particular activity permutation (activity list). When a specified number of ***individuals*** are generated, together they form the initial population. The **2PGA** then uses SSGS to generate a schedule for each ***individual*** and takes the schedules' ***make-span*** as ***individuals'*** fitness measure. An ***individual*** is the encoding of a schedule, it is characterized by a particular permutation of activities, which can be considered as an individual's DNA, if this ***individual*** is chosen as a parent, part of its DNA will be passed to the next generation.

## **Forword-Backward Improvement (a.k.a. double justification)**

Forward scheduling is a procedure, in which activities are scheduled as early as possible subject to their natural precedence and resource capacity constraints. In contrast, backward scheduling works on the same activities and resource capacity constraints, but activity precedence relations are reversed, i.e., successors become predecessors, and vice versa. The backward scheduling can be thought as starting at a given project finish time and scheduling in backward direction. The effect of the backward scheduling is to make the activities in a project start/finish as late as possible for the given project finish time.

Studies have shown that the backward scheduling can often improve schedules (in terms of make-span) generated by forward scheduling [24].

In the **2PGA**, both forward and backward scheduling perform the same type of crossover, mutation, and selection operations. The backward scheduling is based on the population produced by the forward scheduling, and vice versa, the iteration of alternating precedence directions is as follows:

0. **For a given population, repeat:**
1. **Selection, crossover, and mutation to produce the next population based on forward precedence, then reverse the precedence relations, do step 2,**
2. **Selection, crossover, and mutation to produce the next population based on reversed precedence, then reverse the precedence relations, do step 1.**

## **Selection**

In a Genetic Algorithm (GA), **selection** imitates the natural selection in evolution. Its purpose is to select ***individuals*** (candidate solutions) from the current population as parents for the next generation. The fundamental idea is the **Survival of the Fittest**. Solutions with better fitness should have a greater chance to be selected and pass their DNA (solution characteristics) to the next generation of individuals.

By biased on good solutions, selection guides the GA explore the most promising regions of the solution space, incrementally improving the overall quality of the population.

The search focus in solution space is influenced by adjusting the selection pressure.

**Strong selection pressure**, i.e., highly biased on the current best individuals, leads to search intensification in the current neighborhood of solution space, facilitating faster convergence but may be trapped in local optima.

**Weak selection pressure**, i.e., less biased on the current best individuals, leads to search diversification in the solution space, facilitating exploring a wider range of solutions, while leading to slow convergence but with better capability to escape local traps.

The selection pressure can be adjusted. For instance, in the **Tournament Selection** method, $\boldsymbol{k}$ individuals are randomly chosen from the current population, and the individual with the highest fitness is selected as a parent. For this method, increasing parameter $\boldsymbol{k}$ makes **selection pressure** stronger**.**

The GA incrementally improve solutions through the selection from populations based the **survival of the fittest** principle. However, the best current solutions may also be local traps that prevent the GA from searching wider areas. Therefore, the key to success is the **balance** between the intensification and diversification of search in solution space.

We address the **balance** issue using a 2-phase GA (**2PGA**) approach. The only difference between the two phases lies in the **selection**. The basic selection method in **2PGA** is described as follows.

Assume for a given population denoted as set ***POP,*** $N_p$parents are to be selected for the next generation. For both phases, the population is partitioned into two sets: the first set contains individuals with the best fitness (make-span) in the population, the set of such individuals is denoted as ***F*** with size $N_F$. The second set is ***POP**F.***

***Phase 1.*** All the individuals in ***F*** are selected as the parents, and remaining $(N_p - N_F)$ parents are selected from the individuals in ***POP**F***.

***Phase 2.*** All the individuals in ***F*** are excluded from the parent pool. And all $N_p$ parents are selected from the individuals in ***POP**F.***

For both phases, the parent selection is biased on the ***individuals*** with better fitness in ***POP**F***, for this purpose, any relevant selection methods can be used (e.g., Roulette Wheel, Tournament, Rank-Based, Elitism, etc.).

## 2PGA Procedure

The goal of the **2PGA** is to find an ***individual*** (an individual is simply a permutation of all activities), when decoded as a schedule, has the shortest ***make-span***.

The **2PGA** uses: two-point crossover, mutation as a combination of one-activity-move, activity-exchange, group-activity-move, SSGS schedule generation scheme, forward backward improvement, and the 2-phase selection method, as described in the previous sections. A general description of **2PGA** is given as follows.

## 2PGA Parameters

The parameters include: the number of individuals in population; the number of parents for each generation; the numbers of iterations and the sizes of the ***F*** for Phase 1 and phase 2.

The parameters are problem size dependent and can be dynamically adjusted to 1) expand or narrow the range of search and 2) change the **balance** between strong and weak selection pressures (i.e., the balance of search intensification and diversification).

A ***candidate solution list*** is used to stores current best solutions (individuals), the individuals in the list also serve as high fitness parents when the population fitness deteriorates. The list also contains the final solution.

## Phase 1/2 Iteration

In the **2PGA**, the only difference between phase 1 and phase 2 is the parents selection methods. The phase 1/2 iteration is the following

**0. Start from a current population**

**1. Perform Phase 1/2 iteration with the specified number of times**

- **1.1 select parents from the current population using phase 1/2 method.**
- **1.2 crossover parents to generate a new population and set it as the current population.**
- **1.3 perform mutation on the current population.**
- **1.4 compute fitness of the current population based on SSGS generated schedules.**
- **1.5 update the candidate solution list.**
- **1.6 in case individuals have forward activity precedence, reverse the precedence, otherwise, reverse backward precedence to forward precedence, repeat steps 1.1~1.5.**

**2PGA evolution process**

1. **Set parameters, initialize the candidate solution list**
2. **Generate the initial population as the current population, compute fitness of the population**
3. **2PGA Main Iteration**
   - **3.1 Phase 1 iteration**
   - **3.2 Phase 2 iteration**
   - **3.3 If termination condition is satisfied, the current best solution is the final solution, otherwise, repeat step 3.1~3.3.**

## 4 Computation Results

**PSPLIB** is a widely used standard benchmark library for testing **RCPSP** algorithms. It has four datasets J30, J60, J90, and J120, containing **RCPSP** problems with 30, 60, 90, and 120 activities, respectively, the difficulty increases with the number of activities. Each dataset also contains the current best solutions.

The **2PGA** is tested on all the four datasets. At the time of testing, it had reproduced all the current best solutions to the four datasets. Furthermore, it also improved a number of current solutions for J120. At the time of this writing, the **2PGA** accounts for 90 best current heuristic solutions for J120 dataset, as shown in j120hrs.sm in **PSPLIB** [17], which are also shown in Table 1.

**Table 1**. The current best J120 heuristic solutions produced by **2PGA**

| RCPSP No. | Make Span | RCPSP No. | Make Span | RCPSP No. | Make Span | RCPSP No. | Make Span | RCPSP No. | Make Span |
|---|---|---|---|---|---|---|---|---|---|
| 6_5 | 125 | 14_1 | 85 | 19_5 | 103 | 37_6 | 163 | 52_8 | 158 |
| 7_8 | 97 | 14_10 | 81 | 26_10 | 183 | 37_8 | 178 | 57_1 | 185 |
| 7_9 | 89 | 16_3 | 234 | 27_8 | 140 | 38_2 | 124 | 57_2 | 161 |
| 8_4 | 94 | 16_4 | 200 | 31_9 | 189 | 38_3 | 155 | 57_3 | 184 |
| 11_8 | 162 | 17_1 | 140 | 32_2 | 131 | 38_4 | 140 | 57_5 | 179 |
| 12_2 | 117 | 17_2 | 123 | 32_3 | 145 | 38_6 | 122 | 57_7 | 166 |
| 12_3 | 136 | 17_3 | 108 | 32_5 | 138 | 38_8 | 125 | 57_8 | 162 |
| 12_4 | 125 | 17_4 | 120 | 32_6 | 128 | 39_2 | 108 | 57_9 | 167 |
| 12_5 | 162 | 17_5 | 129 | 32_7 | 122 | 39_3 | 111 | 57_10 | 167 |
| 12_6 | 121 | 17_6 | 136 | 32_8 | 135 | 39_4 | 98 | 58_1 | 141 |
| 12_8 | 119 | 17_7 | 146 | 32_9 | 127 | 39_7 | 104 | 58_2 | 126 |
| 12_9 | 105 | 17_8 | 127 | 33_3 | 107 | 46_5 | 149 | 58_3 | 120 |
| 12_10 | 143 | 17_9 | 134 | 33_4 | 112 | 47_5 | 127 | 58_4 | 145 |
| 13_4 | 112 | 17_10 | 134 | 33_5 | 142 | 47_6 | 137 | 58_6 | 140 |
| 13_5 | 91 | 18_1 | 138 | 33_8 | 111 | 47_7 | 118 | 58_7 | 147 |
| 13_6 | 99 | 18_6 | 134 | 33_10 | 106 | 51_1 | 206 | 58_8 | 132 |
| 13_9 | 85 | 18_9 | 91 | 37_1 | 145 | 52_5 | 168 | 58_9 | 130 |
| 13_10 | 92 | 19_4 | 106 | 37_3 | 139 | 52_6 | 196 | 58_10 | 131 |

## 5 Concluding Remarks

Genetic Algorithm (**GA**) is based on the **survival of the fittest** principle. The parent selection for the next generation is biased on the individuals with better fitness. In general **GA** implementations, diverse individuals are generated through the randomization in **selection**, **crossover**, and **mutation**.

In contrast, the **2PGA** explicitly alternates **the inclusion and exclusion** of the fittest individuals in the **selection**, while still using randomization for **crossover**, **mutation,** and also for the completion of **selection**. In so doing, the balance between search intensification and diversification can be adjusted. The computation results have shown the effectiveness of this approach.

After the exclusion of the fittest individuals from parent pools, while populations become more diversified, the fitness may suffer. As a result, the search may proceed in a worsening direction. However, for problems with multiple local optima, allowing going worse sometimes is the only way to navigate through the complex terrain and escape local traps. In case of the population deteriorating, the stored current best solutions can be used to recover the population fitness. Note that similar approach is also used by other optimization methods.

The **Tabu Search [9, 10]** is an improved local search with the capability to escape local optima. It uses a tabu list to forbid recently visited solutions or moves. In doing so, it may force the search to move in a worsening direction. But it is such moves that make the **Tabu Search** work. The **simulated annealing** also allows worsening moves though with decreased probability.

In the context of **nonlinear programming**, the **Watchdog** method [14] is used to reach the global optimum. The method allows the object function to become worse in the line search steps. Its **Watchdog** element conducts periodic checks. After a certain number of worsening steps, the algorithm then uses backtrack or reset to bring the search back on track.

Note also that the **variation of selection pressures** during the **GA** iterations can be implemented in different ways, e.g., not limited to two phases or changing the **selection pressures** gradually. The **2PGA** is just one of the implementations.

In summary, the **2PGA** represents a useful way for balancing search intensification and diversification in **GA.** The computation results have shown its effectiveness for solving **RCPSP.** And as a general approach, the **2PGA** may also be applied to other optimization problems.